%% file: arxiv_new.tex
\documentclass{arxiv_new}

\usepackage{graphicx,wrapfig}
\graphicspath{{./images/}}

\usepackage{url}
\usepackage{latexsym}

\usepackage{xcolor}
\usepackage[utf8]{inputenc}

\usepackage{listings}
\usepackage{booktabs}

\usepackage{multirow}

\issue{6}{2}{2020}

\title{ETC-NLG: End-to-end Topic-Conditioned Natural Language Generation}

\runningtitle{ETC-NLG: End-to-end Topic-Conditioned Natural Language Generation}

\runningauthor{Carbone and Sarti}

\begin{document}

\author{Ginevra Carbone\thanks{Dept. of Mathematics and Geoscience - Via Weiss 2, 34128 Trieste, Italy. E-mail:~\texttt{ginevra.carbone@phd.units.it}}}
\affil{Universit\`a degli Studi di Trieste}

\author{Gabriele Sarti\thanks{Dept. of Mathematics and Geoscience - Via Weiss 2, 34128 Trieste, Italy. E-mail:~\texttt{gabriele.sarti996@gmail.com}}}
\affil{Universit\`a degli Studi di Trieste, SISSA}

\maketitle

\newcommand{\gc}[1]{{\color{blue}[GC: #1]}}
\newcommand{\gs}[1]{{\color{red}[GS: #1]}}

\maketitle

\input{ijcol/sections/Abstract}
\input{ijcol/sections/Introduction}

\input{ijcol/sections/Background}

\input{ijcol/sections/Method}
\input{ijcol/sections/Experiments}
\input{ijcol/sections/Conclusions}

\clearpage

\bibliographystyle{ijcol/fullname}
\typeout{}
\bibliography{bibliography}

\end{document}

%% file: ijcol/sections/Abstract.tex
\begin{abstract}
    
Plug-and-play language models (PPLMs) enable topic-conditioned natural language generation by combining large pre-trained generators with attribute models to steer the predicted token distribution towards selected topics. Despite their efficiency, the large amounts of labeled texts required by PPLMs to effectively balance generation fluency and proper conditioning make them unsuitable to low-resource scenarios. We present \textbf{ETC-NLG}, an approach leveraging topic modeling annotations to produce \textbf{E}nd-to-end \textbf{T}opic-\textbf{C}onditioned \textbf{N}atural \textbf{L}anguage \textbf{G}enerations over emergent topics in unlabeled document collections. We test our method's effectiveness in a low-resource setting for Italian and perform a comparative evaluation of ETC-NLG for Italian and English using a parallel corpus. Finally, we propose an evaluation method to automatically estimate the conditioning effectiveness from generated utterances.

\end{abstract}

%% file: ijcol/sections/Introduction.tex
\section{Introduction}

Pre-trained neural language models can be used for natural language generation (NLG) by autoregressively sampling the most probable token from the learned vocabulary distribution given previous context. Among the most effective autoregressive models, GPT variants~\cite{gpt,gpt2,gpt3} follow the two-step process originally introduced by ULMFiT~\cite{ulmfit}, combining an unsupervised pretraining over massive textual resources with a task-specific fine-tuning to solve a variety of different problems, ranging from machine translation to text summarization. Despite their effectiveness for standard NLG, GPT-like models are still mostly inefficient for advanced forms of NLG, such as topic and sentiment-conditioned generation, requiring fine-tuning with attribute-specific data or even radically changing the model's architecture~\cite{keskarCTRL2019} to allow for better control over generated outputs. \textit{Plug-and-play language models (PPLMs)}~\cite{plugandplay} were recently introduced to counter this tendency, allowing users to efficiently generate controlled text by combining a standard pre-trained language model generator with a discriminator network that learns to differentiate attributes and to steer text generation towards the selected conditioning. 

While simple BoW models have been demonstrated to perform effective conditioning in the context of sentiment analysis \cite{li-etal-2018-delete}, differentiating abstract thematic categories often requires training more sophisticated discriminators, which capture semantic relations between entities from distributed representations.
For this reason, PPLMs need large quantities of annotated documents to train discriminators that are capable of successfully steering the generation process. This fact makes them mostly unsuitable for low-resource domains and languages where such annotations are often unavailable.
To address this weakness, we propose \textbf{ETC-NLG}, an approach leveraging the efficiency of PPLMs and the effectiveness of contextual~\cite{bianchi2020crosslingual} and combined~\cite{bianchi2020pretraining} topic models to enable an \textbf{E}nd-to-end \textbf{T}opic-\textbf{C}onditioned \textbf{N}atural \textbf{L}anguage \textbf{G}eneration from unlabeled document collections.\footnote{Code and materials available at \url{https://github.com/gsarti/ETC-NLG}.} Our approach follows the intuition that exploiting data-driven topics (i.e. categories extracted from unlabeled corpora) is especially beneficial in presence of large collections of text for which the manual annotation process would require significant resources. For example, ETC-NLG could be leveraged in a multi-step setting alongside a large set of unlabeled product reviews to generate category-dependent synthetic examples, continually improving the performances of a classifier that can also be used as category discriminator during generation. Our experiments start from the evaluation of ETC-NLG on the Svevo Corpus~\cite{svevo-analysis,svevo-analysis2}, a topic-annotated Italian epistolary corpus containing archaic and dialectal terms. We then compare the effectiveness of Italian neural language models used by ETC-NLG with their widely-used English counterparts and test their topic modeling and conditioned generation performances on a portion of the EuroParl Italian-English parallel corpus~\cite{europarl}. Lastly, we assess the quality of generated utterances and propose an automatic method to evaluate their conformity to the selected conditioning topic.

%% file: ijcol/sections/Background.tex
\section{Background}

\paragraph{Topic-conditioned NLG} The problem of \textit{topic-conditioned NLG} (also referred to as controlled NLG) involves the generation of a sequence of tokens $X_a = (x_1, x_2, \dots x_n)$ matching a specific topic or category $a$ defined beforehand. The most common approaches adopted to achieve this result requires a full-fledged training of NLG models on specific sets of text categories. For example, Ziegler et al.~\shortcite{Ziegler2019FineTuningLM} fine-tune a GPT-2 model on positive-negative categories using reinforcement learning, while the CTRL transformer-based model by Keskar et al.~\shortcite{keskarCTRL2019} is pre-trained over a fixed set of over 50 control codes representing different textual categories. While this approaches are highly effective in producing fluent and conditioned text because of the direct maximization of $p(x|a)$ (i.e., the probability of each generated token given the selected attribute $a$), they suffer from a lack of flexibility and expensive retraining procedure.  The \emph{Plug-and-Play Language Model}~\cite{plugandplay} (PPLMs) approach used in ETC-NLG and presented later in this section solves these issues by flexibly combining large general-purpose NLG systems with small ad-hoc attribute discriminators, that can be easily retrained in different scenarios.

\paragraph{Topic modeling} Topic models are statistical approaches used to extract latent (i.e. unobserved)  topics occurring in collections of documents. The most well-known approach in the topic modeling domain is the \textit{Latent Dirichlet Allocation}~\cite{lda} (LDA), where a mixture of topics represented by key co-occurring words is associated to each document by means of a generative process. Despite the success of LDA for most common use-cases, recent work in the domain of topic modeling explores the usage of contextual embeddings produced by trained neural language models to augment the performances of standard topic modeling approaches. The semantic information encoded by contextual embeddings was notably shown to be beneficial to improve the coherence across topic-related words~\cite{bianchi2020pretraining}.

\paragraph{Combined and Contextual Topic Models} 

Combined topic models~\cite{bianchi2020pretraining} extend the Neural-ProdLDA~\cite{srivastava2017autoencoding} variational approach by concatenating the input BoW document representation with pre-trained contextual embeddings produced by a neural language model after a pre-training procedure. An inference network maps the resulting vectors to a latent representation, which is then variationally sampled by a second decoder network to reconstruct the document BoW, effectively approximating the standard Dirichlet prior~\cite{lda} with normally-distributed samples. Fully-contextual topic models~\cite{bianchi2020crosslingual} follow the same principle, but language models' contextual embeddings entirely replace the BoW representation. The main advantage of these approaches over classical LDA-like methods for topic modeling is the use of semantically-informed representations, which were shown to improve intra-topic coherence and provide more meaningful classifications. In our work, combined and contextual topic models are used to produce synthetic topic annotations from unlabeled document collections. These annotations are later used to train a discriminator network used as conditioning component in the PPLM approach.

\paragraph{Plug-and-play Language Models} 

Plug-and-play language models flexibly combine large pre-trained language models with cheap discriminators, to perform controlled text generation. These models follow a Bayesian approach by including:
\begin{itemize}
\setlength\itemsep{0.5em}
    \item An \emph{unconditional language model} generator, acting as a prior probability distribution $p(x)$ over text.
    
    \item An \emph{attribute model} discriminator $p(a|x)$, expressing the likelihood of an attribute $a$ given text $x$.
    
    \item The resulting \emph{conditional language model} $p(x|a)$, used for conditional autoregressive generation.
\end{itemize}

This approach is appealing since tiny and cheap attribute models can produce excellent conditional generative models when combined with powerful pre-trained architectures such as GPT-2, without any fine-tuning of the language model on attribute-specific data. We point out that PPLMs can be applied to any transformer-based language model. The performances of PPLMs were shown to be comparable to those of cumbersome fully-conditional language models like CTRL~\cite{keskarCTRL2019} with far less computation. In should be noticed that the expression ``attribute'' in this case refers to an abstract topic, but the same approach could be extended to other classes of attributes, such as positive and negative sentiments.

\paragraph{Steering generation}

The objective of PPLM is that of steering the representation of text toward high log-likelihood values for both the conditional attribute model $p(a|x)$ and the unconditional language model $p(x)$. Maximizing $\log p(a|x)$ ensures that $x$ will more likely possess attribute $a$, while maximizing $p(x)$ guarantees fluency within the generated text.
These updates are restricted to a specific part of the model, which accounts for the past information. Authors rely on the recurrent notation of transformer models, where at each time step the \emph{history matrix} $H_t$ of past key-value pairs \cite{transformers} is available
$$
H_t = [(K_{0:t}^{(1)},V_{0:t}^{(1)}),\ldots,(K_{0:t}^{(l)},V_{0:t}^{(l)})].
$$
In this setting, the updates are performed in the continuous space of the hidden representations, with the aim of gradually reinterpreting the past information given by $H_t$.

The attribute model $p(a|x)=p(a|H_t+\Delta H_t)$ is updated in the direction of its gradient 
$$\nabla_{\Delta H_t} \log p(a|H_t+\Delta H_t)$$ with respect to the history matrix, which is null-initialized. The perturbed history matrix produces a perturbed language model $\tilde{p}_{t+1}$ at the next time step, which would otherwise be a model $p_{t+1}$ depending on the unperturbed previous history. This process is repeated multiple times and, after each iteration, $\Delta H_t$ is further updated in favour of high $p(x)$ regions, by minimizing the KL divergence between the perturbed model $\tilde{p}_{t+1}$ and the unperturbed model $p_{t+1}$. Additionally, the strategy of \emph{post-norm fusion similarity} \cite{plugandplay} is applied at each step with the aim of preserving text fluency.

\paragraph{Sampling from the conditional LM}

Starting from a desired attribute $a$ and a prefix sentence\footnote{We could also avoid setting the beginning of the sentence by using \lstinline{|<endoftext>|} as a prefix.}, the sampling steps of PPLM can be summarized as follows:
\begin{itemize}
    \item The first step is a forward pass through the language model to compute the likelihood of the attribute $p(a|x)$;
    \item The second step is a backward pass that updates the latent representation of text, using gradients from the attribute model
    $$H_t \longleftarrow H_t+\Delta H_t,$$
    as described in the previous section;
    \item In the third step next token $x_{t+1}$ is sampled form the perturbed distribution $\tilde{p}_{t+1}$, which depends on the current token $x_t$ and the new latent representation $\tilde{H}_t=H_t + \Delta H_t$.
    
\end{itemize}

%% file: ijcol/sections/Method.tex
\section{Methodology}

\begin{figure}[htbp]
\begin{center}
\includegraphics[width=0.83\textwidth]{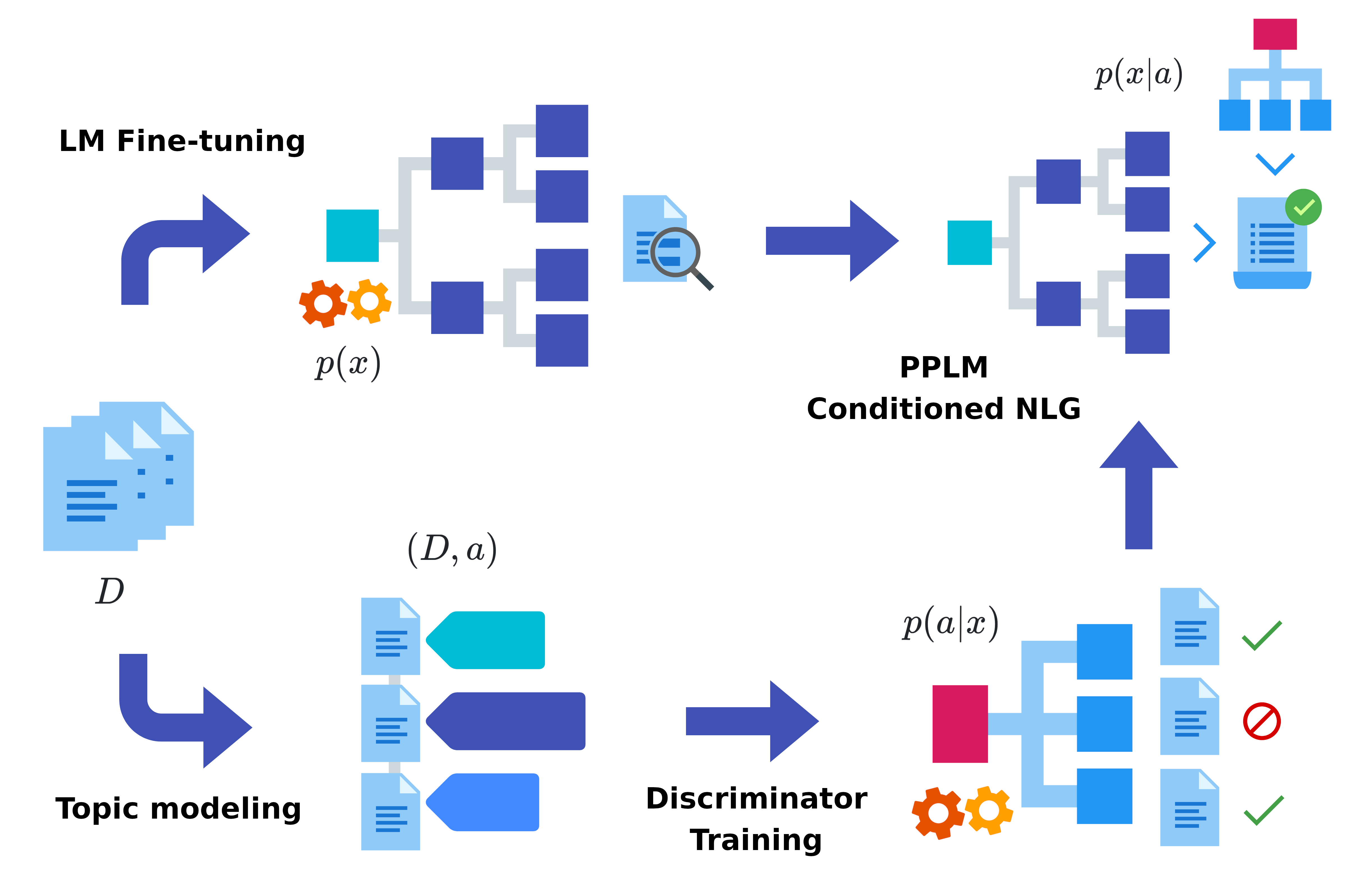}
\caption{A visual representation of the End-to-end Topic-Conditioned Natural Language Generation (ETC-NLG) pipeline. An unlabeled document collection is used to train a language model (top) and a topic model (bottom). Automatic topic annotations are used to condition the generation of a PPLM through a discriminator network.
}
\label{fig:pipeline}
\end{center}
\end{figure}

Figure~\ref{fig:pipeline} offers a visual representation of our approach building upon the PPLM architecture. The pipeline starts from an unlabeled document collection $D$  (left) and fine-tunes a neural language model generator to adapt its predicted distribution over the vocabulary to the current setting, producing an unconditional language model $p(x)$ (top-left). Besides, it performs an automatic topic modeling over the documents $D$, using either combined or contextual topic models. Trained topic models are subsequently used to annotate each document in $D$ with its most relevant topic, producing a collection of topic-annotated documents $(D,a)$ (bottom-left). Automatic topic annotations are used to train an attribute model discriminator $p(a|x)$ that predicts document topics given their contextual embedding representations (bottom-right). Finally, the two networks merge into a PPLM conditional language model $p(x|a)$, for the generation of topic-conditioned utterances (top-right). Language and topic modeling steps can be performed in parallel to reduce the overall training time, taking advantage of the independence between generative and discriminative components of ETC-NLG.

While this approach is advantageous when dealing with insufficient labeled data and low-resource scenarios, since topic labels are inferred, the production of meaningful sentences heavily relies on topic modeling quality. We discuss this perspective after describing the experimental results of Section~\ref{sec:experiments}.
Compared to a classical language modeling fine-tuning followed by topic induction, ETC-NLG is more computationally efficient due to his composable nature. Most importantly, it allows the user to manually set the hyperparameters which are responsible for quality of the final outcome, e.g. the strenght of conditioning or the fluidity of text. This aspect is clearly beneficial when different degree of conditioning may be important in the generated sequences.

%% file: ijcol/sections/Experiments.tex
\section{Experimental Results}
\label{sec:experiments}

Our experimental objectives are three-fold: first, we test ETC-NLG on the Italian subset of the epistolary corpus of Italo Svevo~\cite{svevo-analysis}, a famous Italian author of the early 20th century, to quantify the impact of dialectal and archaic expressions on the quality of generated sentences. Secondly, we compare the performances of ETC-NLG on Italian and English by leveraging a portion of the European Parliament Proceedings (EuroParl) parallel corpus~\cite{europarl}. Finally, we perform an empirical evaluation of the obtained results and present an intrinsic evaluation method based on topic modeling quality over conditionally-generated texts.

\subsection{Data} The Svevo Corpus contains 5419 sequences ranging from few words to multiple sentences. Each sequence is annotated with one or more topics by two domain experts using a set of five main topics (family, literature, work, travel, health) and five sub-topics that were found during a previous analysis of the corpus~\cite{svevo-analysis2}. We aggregate most of the sub-topics with main topics to reduce sparsity and obtain a final set of documents annotated by 6 topics: \textit{family} (1669 seq.), \textit{wife} (1298 seq.), \textit{travel} (821 seq.), \textit{health} (552 seq.), \textit{literature} (544 seq.) and \textit{work} (535 seq.). The EuroParl corpus does not contain topic annotations and comprises almost 2 million parallel sentences collected from parallel Italian-English proceedings of the European Parliament. We only select the first 50'000 sentences for our modeling experiments.

\begin{wrapfigure}[23]{r}{5.6cm}
    \includegraphics[width=\linewidth]{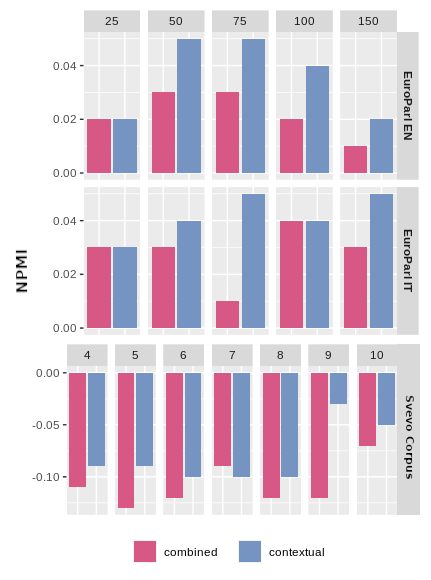}
    \caption{
    NPMI scores for contextual and combined topic models over the three corpora with variable topic counts. Higher scores correspond to higher relatedness between topic words.}
    \label{fig:npmi}
\end{wrapfigure}

Table \ref{tab:example_labels} reports examples of Svevo corpus gold annotations (top part), which are manually annotated by expert curators, and English Europarl annotations (bottom part), produced using a contextual topic model.  Sentences can have multiple labels with possibly overlapping meanings, causing a higher uncertainty in predictions for the resulting discriminator model. Therefore, the quality of topic annotations plays a fundamental role. The interpretation of topics is often straightforward for the Europarl corpus, while it may require significant domain-specific knowledge in the case of the Svevo corpus. Readers are referred to~\cite{svevo-analysis,svevo-analysis2} for additional details on the latter.

\begin{table}[ht!]
\caption{Examples from the Svevo Corpus with \textcolor{purple}{gold} 
annotations (top), with each keyword representing a different topic, and from the English Europarl corpus with \textcolor{purple}{contextual} annotations (bottom), where the top 5 topic-related keywords from the prevalent topic extracted through topic modeling are presented.}
\begin{small}
\begin{tabular}{p{\textwidth}}
\toprule
\textbf{Svevo corpus with gold labels}
\\\toprule
Nel corso di questo mese sarò probabilmente a Parigi. Verrò a salutarla. Porterò con me gli articoli più importanti (pochissimi) che mi furono dedicati. Già non credo ch'Ella abbia premura. Per il momento Ella ha molte altre cose cui pensare prima che all'articolo da dedicare a me. Ha visto l'articolo di Marcel Thiébaut nella « Revue de Paris» del 15 Novembre? Per dire il vero l'articolo del «Baretti» non mi piacque molto. Nell'ultimo «Convegno» di Milano c'è un articolo di Sergio Solmi pieno di belle osservazioni. \textcolor{purple}{[Travel, Literature]}\\
\midrule
E quando il momento è brutto — ed è anzi il momento brutto quasi sempre perché è l'ora in cui si sentirebbe più il bisogno di una lieta compagnia, di un appoggio, di un incoraggiamento —, scrivendoti cicco quasi sempre e dovresti ancora essere contenta che tante volte il pensiero a te mi mitiga. Iersera presi bromuro e una di quelle tue Purgen e dormii bene. \textcolor{purple}{[Wife, Health]}
\\
\midrule
Ma io desidero vivamente ch'Ella conosca anche l'originale  del Zeno. Non soltanto perché è là cosa che — checché ne dicano i malevoli — scrissi meglio, ma anche perché per volere del Gallimard la traduzione fu falcidiata di non meno di 100 pagine. Come se in francese non esistessero dei romanzi più lunghi del mio. Fui un po' loquace, è vero, ma è doloroso, specialmente per un loquace, di sentirsi tagliare la parola. Una ferita che — secondo il Freud — è esposta alla soppressione patologica. \textcolor{purple}{[Literature]}\\
\toprule
\textbf{Europarl corpus with contextual labels}
\\\toprule
I believe that the principle of relative stability is a fundamental legal principle of the common fisheries policy and a proposal to subvert it would be legally inadmissible. I want to know whether one can raise an objection of that kind to what is merely a report, not a legislative proposal, and whether that is something I can competently do on Thursday. \textcolor{purple}{[Member, State, Procedure, Provision, Budget]}\\
\midrule
Madam President, the presentation of the Prodi Commission' s political programme for the whole legislature was initially a proposal by the Group of the Party of European Socialists which was unanimously approved by the Conference of Presidents in September and which was also explicitly accepted by President Prodi, who reiterated his commitment in his inaugural speech. \textcolor{purple}{[Committee, Group, Amendment, Behalf, Report]}\\
\midrule
Prevention has to be our answer to disasters of this kind and this draft directive is an important step towards well-trained safety advisers being available, so that the right action is taken in good time. \textcolor{purple}{[Safety, Food, Waste, Product, Animal]}\\
\bottomrule
\end{tabular}
\end{small}

\label{tab:example_labels}
\end{table}

\begin{table}[t!]
\caption{Result of topic modeling evaluation using topic diversity ($\alpha$), inverted RBO ($\rho$) and NPMI ($\tau$) metrics. \textbf{Bold} models were selected for conditioned generation experiments.}
    \centering
    \begin{small}
    \begin{tabular}{lcccccc}
    \toprule
    & \multicolumn{3}{c}{\textbf{Contextual}} & \multicolumn{3}{c}{\textbf{Combined}} \\
    \cmidrule(lr){2-4}
    \cmidrule(lr){5-7}
    & $\alpha$ & $\rho$ & $\tau$ & $\alpha$ & $\rho$ & $\tau$ \\
    \midrule
    Svevo Corpus-4  & .93 & .98 & -.09 & .99 & 1.00 & -.11 \\
    Svevo Corpus-5  & .85 & .97 & -.09 & .97 & 1.00 & -.13 \\
    \textbf{Svevo Corpus-6}  & .72 & .92 & -.10 & .91 & .98 & -.12 \\
    Svevo Corpus-7  & .71 & .94 & -.10 & .91 & 1.00 & -.09 \\
    Svevo Corpus-8  & .73 & .93 & -.10 & .79 & .95 & -.12 \\
    Svevo Corpus-9  & .69 & .94 & -.03 & .77 & .96 & -.12 \\
    Svevo Corpus-10 & .64 & .91 & -.05 & .78 & .96 & -.07 \\
    \midrule
    EuroParl-IT-25  & .79 & 1.00 & .03 & .67 & .99 & .03 \\
    EuroParl-IT-50  & .54 & .99 & .04 & .41 & .98 & .03 \\
    \textbf{EuroParl-IT-75}  & .42 & .99 & .05 & .24 & .96 & .01 \\
    EuroParl-IT-100 & .33 & .98 & .04 & .23 & .96 & .04 \\
    EuroParl-IT-150 & .23 & .98 & .05 & .17 & .96 & .03 \\
    \midrule
    EuroParl-EN-25  & .82 & 1.00 & .02 & .72 & .99 & .02 \\
    EuroParl-EN-50  & .56 & .99 & .05 & .44 & .99 & .03 \\
    \textbf{EuroParl-EN-75}  & .43 & .99 & .05 & .32 & .98 & .03 \\
    EuroParl-EN-100 & .35 & .99 & .04 & .15 & .94 & .02 \\
    EuroParl-EN-150 & .16 & .96 & .02 & .13 & .94 & .01 \\
    \bottomrule
    \end{tabular}
    \end{small}
    
    \label{tab:tm_results}
\end{table}

\subsection{Topic Modeling}

We test both combined and contextual topic modeling approaches using RoBERTa~\cite{liu2019roberta}, a widely-known improvement over the BERT encoder~\cite{devlin-etal-2019-bert}, and UmBERTo~\cite{umberto}, a RoBERTa-based encoding language model trained on crawled Italian web data, producing respectively English and Italian contextual representations. We leverage the base variants of both models available through the HuggingFace Transformers framework~\cite{Wolf2019HuggingFacesTS}. Contextual embeddings are sampled either alone or alongside bag-of-words representations in a variational framework to improve topic coherence.

Given the different sizes of tested corpora, we evaluate combined and contextual models' performances by varying the number of topics between 3 and 10 for the Svevo corpus and between 25 and 150 for the EuroParl corpora. We use three metrics capturing topic coherence and diversity\footnote{See works by Bianchi et al.~\cite{bianchi2020crosslingual,bianchi2020pretraining} for specifications on models and metrics.}: i) Normalized Pointwise Mutual Information (NPMI, $\tau$), measuring the relatedness of top-10 topic words given the empirical frequency of corpus words; ii) Topic Coherence ($\alpha$), i.e., the average of pairwise similarities between word embeddings of top-25 topic words across topics for a semantic perspective to topic coherence; and iii) Rank-Biased Overlap diversity (Inverted RBO, $\rho$), a measure of disjointedness between topics weighted on word rankings.
Figure~\ref{fig:npmi} reports the NPMI scores for combined and contextual models over topic ranges for the three corpora. Topics generated by the contextual model are generally more coherent across all topic counts, with EuroParl topics being more coherent than those of the Svevo corpus for both Italian and English languages. 

Table~\ref{tab:tm_results} reports the whole set of evaluated metrics for all models on all corpora. From the results, we can see a clear trade-off between the number of selected topics and the quality of $\alpha$ and $\rho$ metrics, while NPMI does not appear to be affected by topic counts. For our generation experiments, we choose the 6-topics contextual model for the Svevo corpus to match the number of gold topic labels empirically set by human annotators, and the 75-topic contextual models for both Italian and English EuroParl, given their strong performances in both topic coherence and NPMI. The upcoming sections assume that those are the only models used to produce annotations for training the discriminators unless otherwise mentioned.

\subsection{Conditional Text Generation}

\begin{table}[t!]
\caption{A contextual topic model is used to produce labels for training the contextual discriminator on each corpus. No gold labels are available for the EuroParl corpus.}
\centering
\begin{tabular}{c||c|c|c}
    \toprule
        \def\arraystretch{1}\begin{tabular}{@{}c@{}} \textbf{Test} \\ \textbf{performances} \end{tabular} & \def\arraystretch{1}\begin{tabular}{@{}c@{}} Svevo \\ Corpus \end{tabular} & \def\arraystretch{1}\begin{tabular}{@{}c@{}} EuroParl \\ IT \end{tabular} & \def\arraystretch{1}\begin{tabular}{@{}c@{}} EuroParl \\ EN \end{tabular} \\
    \hline
    \hline
        \def\arraystretch{1}\begin{tabular}{@{}c@{}} Fine-tuned \\ LM  perplexity \end{tabular} &
         $37.00$ & $14.04$ & $6.80$\\
    \hline
        \def\arraystretch{1}\begin{tabular}{@{}c@{}} Discriminator\\ test accuracy \\ (Gold) \end{tabular} &
         $62\%$ & - & -\\
    \hline
        \def\arraystretch{1}\begin{tabular}{@{}c@{}} Discriminator \\ test accuracy \\ (Contextual) \end{tabular} &
         $51\%$ & $95\%$ & $91\%$\\
    \bottomrule
\end{tabular}

\label{table:test_performances}
\end{table}
\begin{table}[t!]
\caption{Training hyperparameters were optimized over Svevo Corpus and reused for EuroParl.}
\centering
\begin{tabular}{c|c}
    \toprule
        Base LM & \def\arraystretch{1}\begin{tabular}{@{}c@{}}
        GPT-2 (EN)~\cite{gpt2} \\ 
        GePpeTto (IT)~\cite{geppetto}  
        \end{tabular} \\
    \hline
        LM fine-tuning &  \def\arraystretch{1}\begin{tabular}{@{}c@{}} epochs = $2$\\ max sequence length = $128$  
        \end{tabular} \\
    \hline
        Discriminator & \def\arraystretch{1}\begin{tabular}{@{}c@{}} epochs = $10$ \\ max sequence length = $128$ \end{tabular}
        \\
    \hline
        PPLM & \def\arraystretch{1}\begin{tabular}{@{}c@{}} 
        output sequences length = $60$\\
        iterations = $15$\\ 
        repetition penalty = $1.5$\\
        window length = $0$\\
        horizon length = $5$ \\
        top-k = $10$\\
        step size = $0.3$\\
        $0.9\leq$  gm scale $\leq0.99$\\
        $1.\leq$ temperature $\leq1.5$\\
        \end{tabular}\\
    \bottomrule
\end{tabular}

\label{table:pplm_training}
\end{table}

We use GPT-2 as the PPLM generator for the English EuroParl dataset and its Italian counterpart GePpeTto~\cite{geppetto} for Svevo and Italian EuroParl datasets. In all such cases, we observe that $2$ fine-tuning epochs are enough to obtain adequate models with low perplexity values (Table \ref{table:test_performances}). In particular, we observe that using a low number of iterations for LM fine-tuning, $2$ to $5$, is often optimal. 
Transformers suffer from well-known computational limitations due to the self-attention complexity of $O(n^2)$ w.r.t. a sequence length $n$. For this reason, we cut sentences from the training corpora were at the last punctuation symbol occurring before a chosen maximum sequence length. In our experiments, we used a maximum of $128$ tokens, meaning that LMs were trained on sentences with a variable number of tokens up to this value.

The discriminator consists of a lightweight transformer encoder followed by a dense classification head. It is trained on the ten most frequent topics provided by the automatic annotations of topic models for all corpora. The discriminator is additionally trained on gold labels for Svevo letters, examples of which are reported in Table \ref{tab:example_labels}. Table \ref{table:test_performances} shows the discriminator's best performances in all such scenarios.

\begin{table*}[t!]
\caption{Examples of ETC-NLG topic-conditioned generation from the Svevo corpus using gold labels, with \lstinline{temperature} values between $[1, 1.5]$ and \lstinline{gm\_scale} between $[0.9, 0.99]$. \textcolor{blue}{Blue text} is the conditioning topic, \textbf{bold} represents prefix context.}
\begin{small}
\begin{tabular}{p{\textwidth}}
\toprule
\textcolor{blue}{[Wife]} \textbf{La tua} assenza mi secca. Non ho nulla da dirti e spero che tu potrai
sapere se la lettera sarà spedita in qualche città della mia vita o meno, a Venezia oppure a Milano! \\
\midrule
\textcolor{blue}{[Travel]} \textbf{Un giorno} mi disse che per me sarebbe stato meglio di non venirci a prendere. Se ci fossero stati
due o quattro giorni sarei partito senza di loro e avrei fatto un viaggio simile, in una città più bella della stessa Parigi dove il sole si leva. \\
\midrule
\textcolor{blue}{[Literature]} \textbf{Un giorno} ti scriverò. Non ho scritto che il primo bacio sia stato quello
di Olga. Ho ricevuto la lettera di Letizia e Marco, i due francesi, con le lettere d'ieri. \\
\midrule
\textcolor{blue}{[Work]} \textbf{Se potessi} fare un simile sacrificio
di tempo in ufficio, sarei molto meglio esposto a questo rischio; se tu mi dici di aver bisogno d'operazioni (per esempio la posta) io direi che il lavoro è più facile ma bisogna farlo bene per avere delle idee nuove. \\
\midrule
\textcolor{blue}{[Health]} \textbf{Se potessi} 
avere un po' di riposo per la mia giornata, avrei fatto una grande visita al mio medico di famiglia. Mi disse: «Sai come ti scrivo e mi dici quale cura è quella del tuo male». Poi rispose che non sarebbe stata necessaria l'«iniezione dl» ma solo una o più visite (che può portare allungarsi fino a 2 settimane). \\
\midrule
\textcolor{blue}{[Family]} \textbf{La tua}  lettera non ti scriverà niente. Tanti saluti da Paola, mia cara amica e di tutta la famiglia Villa Veneziani a Trieste; una grande giornata per tutti! \\
\bottomrule
\end{tabular}
\end{small}

\label{tab:svevo_examples}
\end{table*}

The best topic model trained on Svevo letters brought the discriminator to maximum test accuracy of $51\%$, under optimal training settings among those tested. In terms of test accuracy, we observe that a poor discriminator represents a significant bottleneck in generating high-quality sentences from the contextual PPLM. The discriminator trained on gold labels achieves higher test accuracy (Table~\ref{table:test_performances}), showing that manually-annotated sequences are also more separable from a contextual embedding perspective. From Table~\ref{table:test_performances}, we answer our first two experimental objectives by noticing that:

\begin{itemize}
\setlength\itemsep{0.5em}
    \item The dialectal and archaic nature of the Svevo corpus severely cripples both generator and discriminator performances, making conditioned NLG more challenging.
    
    \item Both the generator and discriminator obtain comparable scores on the EuroParl parallel corpora, suggesting that the quality of pre-trained contextual representations of language models (both the one used for language modeling and the one used for topic modeling) is similar across the two languages.
\end{itemize}

We note that two fundamental components for achieving good PPLM performances are a language model with low perplexity over the selected corpus and a discriminator with high test accuracy on corpus annotations. Combining these two properties guarantees a generation of text that is both fluent and adequately contextualized from a human evaluation perspective. 

\begin{table}
\begin{small}
\caption{Examples of ETC-NLG topic-conditioned generations from all corpora using automatically-produced labels. \textcolor{blue}{Blue text} represents conditioning topic, \textbf{bold text} represents prefix context provided for generation, \textcolor{red}{red text} represents topic model prediction during automatic evaluation.}
\begin{tabular}{p{\textwidth}}
\toprule
\textbf{Svevo Corpus}
\\\toprule
\textcolor{blue}{[senilità--devotissimo--joyce--amicare--carissimo]} \textbf{La tua} lettera di ieri, la seguente: «Se ti dicessi che non sei un critico d'arte o uno scrittore ma una persona capace e forte...». Non credo tu ne abbia bisogno. Se mi fosse possibile sarebbe meglio che anche io finisse per scriverti senza avere a cui \textcolor{red}{[senilità--devotissimo--joyce--amicare--carissimo]}\\
\midrule
\textcolor{blue}{[scell--halperson--roncegno--finito--scala]} \textbf{Se potessi} essere a Salso sarei felice di vedermi ancora più vicino. Sto benone. Ho avuto un'idea della mia assenza ma non ho potuto scriverla in nessun modo perché la mia lettera è perduta per il momento e sono stanco solo che dopo aver letto le tue lettere \textcolor{red}{[decembre--tribel--raffreddore--debole--capanna]}\\
\toprule
\textbf{EuroParl EN}
\\\midrule
\textcolor{blue}{[congratulate--excellent--rapporteur--thank--congratulation]} \textbf{It is} therefore important to make the necessary adjustments, especially in relation human rights and democratic principles. There are a number of points which I believe are crucial, but this also affects us all: it must be equal for all our \textcolor{red}{[racism--xenophobia--violence--minority--ethnic]}\\
\midrule
\textcolor{blue}{[state--member--national--small--large]} \textbf{You did} not request an additional explanation. If you look at what the Committee on Economic and Monetary Affairs has to say,  I do agree with your positions. Most of all is that we must bear in mind the need for the single currency - as it was  \textcolor{red}{[market--euro--company--investment--service]} \\
\midrule
\textcolor{blue}{[peace--process--negotiation--agreement--israel]} \textbf{I would} like first to say that I supported the report. We have begun negotiations with Croatia and Macedonia overall, but we will not be able either to join forces in this area or to continue working on it unless there are new conditions. \textcolor{red}{[peace--process--negotiation--agreement--israel]} \\
\toprule
\textbf{EuroParl IT}
\\\midrule
\textcolor{blue}{[acqua--mare--pescare--rifiuto--inquinamento]} \textbf{In questo} modo, si potrà garantire una migliore protezione dell'ambiente e delle risorse a livello mondiale. Per il momento non sono ancora soddisfacenti le previsioni della commissione per l'industria, la ricerca e lo sviluppo tecnologico sulla riduzione del tenore di zolfo nei combustibili liquidi. \textcolor{red}{[acqua--mare--pescare--rifiuto--inquinamento]} \\
\midrule
\textcolor{blue}{[umano--fondamentale--libertà--diritto--carta]} \textbf{Si dovrebbe} invece essere più cauti quando si tratta di stabilire un nesso fra la politica della concorrenza e le condizioni sociali. L'idea che l'Unione europea sia uno strumento per il benessere delle sue popolazioni è in realtà una falsa illusione,  perché non esiste niente al mondo reale. \textcolor{red}{[umano--fondamentale--libertà--diritto--carta]} \\
\midrule
\textcolor{blue}{[produrre--cioccolato--produttore--consumatore--qualità]} \textbf{Si dovrebbe} prestare maggiore attenzione alla prevenzione e al ripristino degli habitat naturali. La biodiversità deve costituire uno dei principali problemi di tutte le politiche comunitarie,  in quanto è l'unico criterio valido per decidere come affrontare i cambiamenti climatici, soprattutto nel settore agricolo; pertanto dobbiamo tenere conto dell'importanza del \textcolor{red}{[acqua--mare--pescare--rifiuto--inquinamento]} \\
\bottomrule
\end{tabular}
\end{small}

\label{tab:cond_sentences}
\end{table}

\subsection{Evaluation}

 We can use the PPLM scheme to produce conditioned sentences after fine-tuning the language model and training the discriminator. We choose four different neutral prefix sentences (see Table \ref{table:prefix_sentences}) for each model and generate three conditioned sentences for all possible combinations of topics and prefixes. We produce a total of 72 sentences on the Svevo Corpus (plus another 72 on gold labels for human evaluation) and 120 sentences each for both EuroParl-IT and EuroParl-EN.

\newcolumntype{P}[1]{>{\centering\arraybackslash}p{#1}}

\begin{table}[!ht]
\caption{Prefix sentences used during PPLM generation. We generate three different sentences for each combination of prefix and conditioning label.}
\centering
\begin{tabular}{c|P{33em}}
    \toprule
        Svevo & \def\arraystretch{1}\begin{tabular}{@{}c@{}} ``Se potessi", ``Io sono", ``La tua" , ``Un giorno"
        \end{tabular} \\
    \hline
        EuroParlIta &  \def\arraystretch{1}\begin{tabular}{@{}c@{}} 
        ``Dato il", ``Si dovrebbe",``Penso che", ``In questo"
        \end{tabular} \\
    \hline
        EuroParlEng & \def\arraystretch{1}\begin{tabular}{@{}c@{}}
        ``It is", ``I would",``You did", ``In this"
        \end{tabular}
        \\
    \bottomrule
\end{tabular}

\label{table:prefix_sentences}
\end{table}

\subsubsection{Human evaluation on gold labels} We start our assessment by manually estimating the quality of conditioned generations. We wish to emphasize that our human evaluation is solely intended as a qualitative assessment performed without external participants' help and is not supported by any statistically significant evaluation. Sentences generated by the PPLM model based on Svevo gold labels show some weaknesses in performing proper conditioning, as expected after the discriminator's poor results. However, they are generally well-formed and coherent from both a morphological and a syntactic perspective. Aiming at stronger conditioning, we perform hyperparameter tuning on the \lstinline{step_size} parameter, controlling the size of a gradient step, the \lstinline{temperature} parameter, inducing a decreased model confidence in its top predictions, 
and the \lstinline{gm_scale} parameter, which accounts for the weight of perturbed probabilities during the history updates. Values chosen for these hyperparameters are reported in Table \ref{table:pplm_training}. Examples of conditioned generation on the Svevo Corpus, presented in Table~\ref{tab:svevo_examples}, show how ETC-NLG can produce meaningful sentences despite the relatively high perplexity achieved by GePpeTto generator on the epistolary corpus.  Additional examples of generated sentences for all corpora are provided in Table~\ref{tab:cond_sentences}.
The generated text is fluid and we observe a strong semantic similarity between predicted and conditioning labels, even when the labels are different. Despite the absence of fine-tuning, sentences generated on the Svevo corpus show an evident shift in language style compared to the original language model, suggesting promising future applications of ETC-NLG in the stylistic transfer domain.

\subsection{Automated evaluation from topic models} We conclude our analysis by proposing a method to automate the assessment of the conditioning intensity achieved in text generation.  We use the same contextual topic models that were initially chosen for labeling the corpora in the ETC-NLG pipeline. In particular, we use them to predict the most likely topic (label) of each conditionally-generated sentence. Then, we judge the quality of topic conditioning by looking at the resulting confusion matrix between the desired and the predicted conditioning topics. We point out that this method can only evaluate the consistency between the generated sentences and the prior topic model, but it cannot estimate their conditioning quality in an absolute sense. 

\begin{figure}
    \centering
    \includegraphics[width=\textwidth]{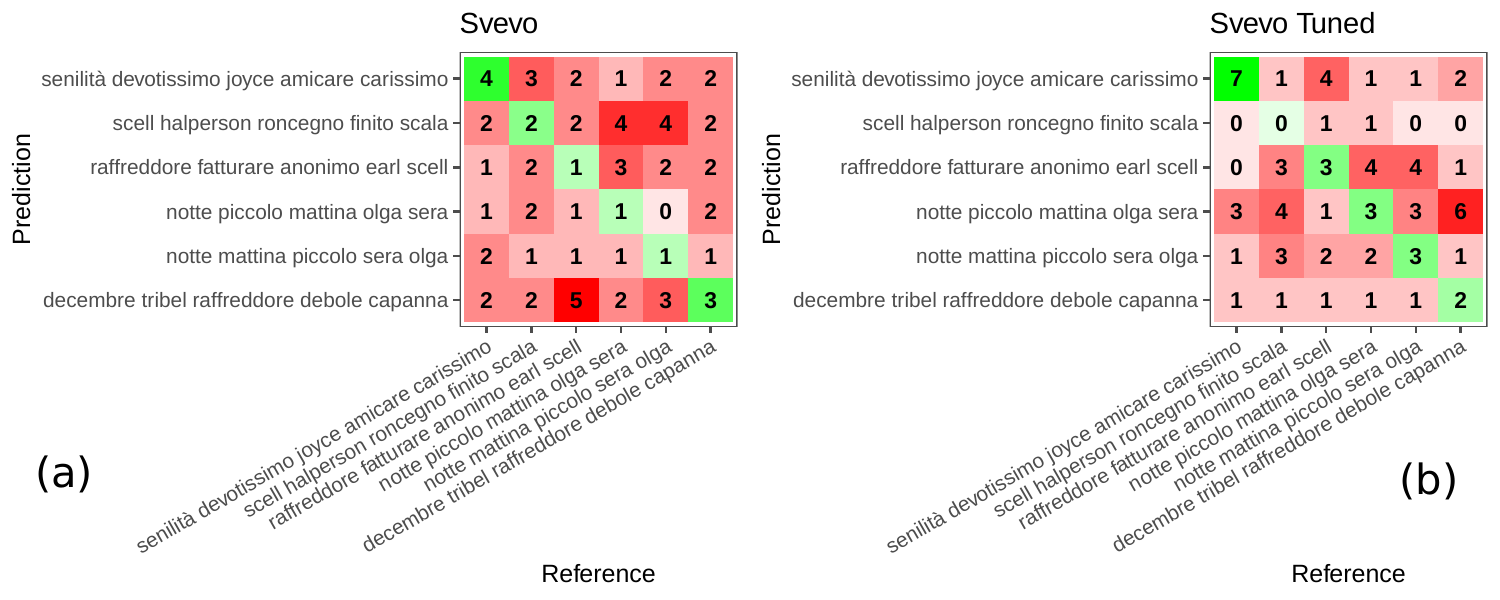}
    \caption{Confusion matrices obtained by predicting the most likely topic from conditionally-generated sentences produced by ETC-NLG on the Svevo corpus. We leverage the same contextualized model that was used to annotate the unlabeled corpora to generate topic predictions and compare those against conditioning labels for sentences generated with weak conditioning parameters \lstinline{step_size}$=0.02$, \lstinline{gm_scale}$=0.9$ in (a) and strong conditioning parameters \lstinline{step_size}$=0.3$, \lstinline{gm_scale}$=0.95$ in (b). Rows represent topic predictions, columns are label references.}
    \label{fig:cm_svevo}
\end{figure}

\begin{figure}[!ht]
\centering
\includegraphics[width=0.85\textwidth]{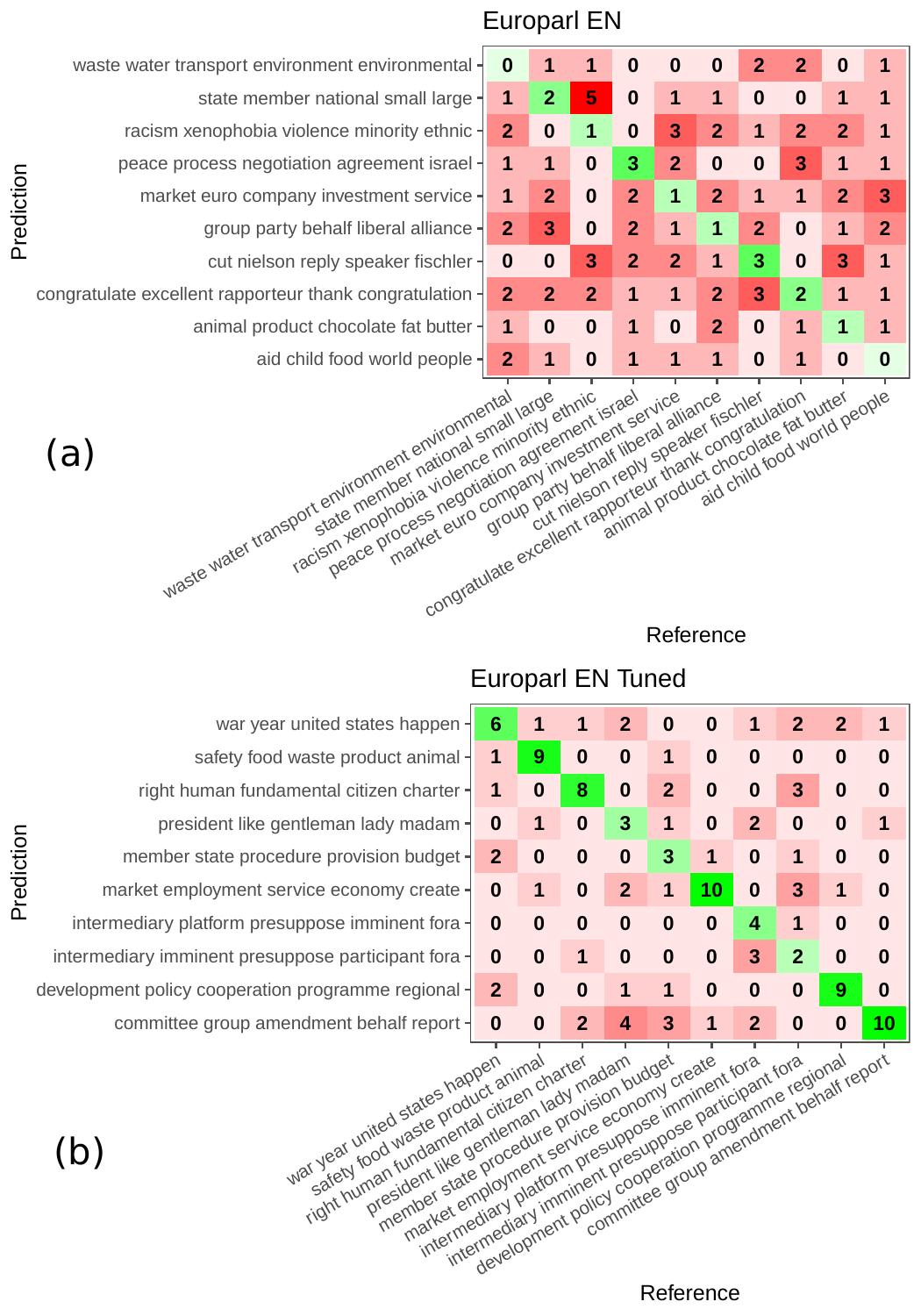}
\caption{Confusion matrices obtained by predicting the most likely topic from conditionally-generated sentences produced by ETC-NLG on the EuroParl English corpus. We leverage the same contextualized model that was used to annotate the unlabeled corpora to generate topic predictions and compare those against conditioning labels for sentences generated with weak conditioning parameters \lstinline{step_size}$=0.02$, \lstinline{gm_scale}$=0.9$ in (a) and strong conditioning parameters \lstinline{step_size}$=0.3$, \lstinline{gm_scale}$=0.95$ in (b). Rows represent topic predictions, columns are label references.}
\label{fig:cm_europarl_eng}
\end{figure}

\begin{figure}[!ht]
\centering
\includegraphics[width=0.85\textwidth]{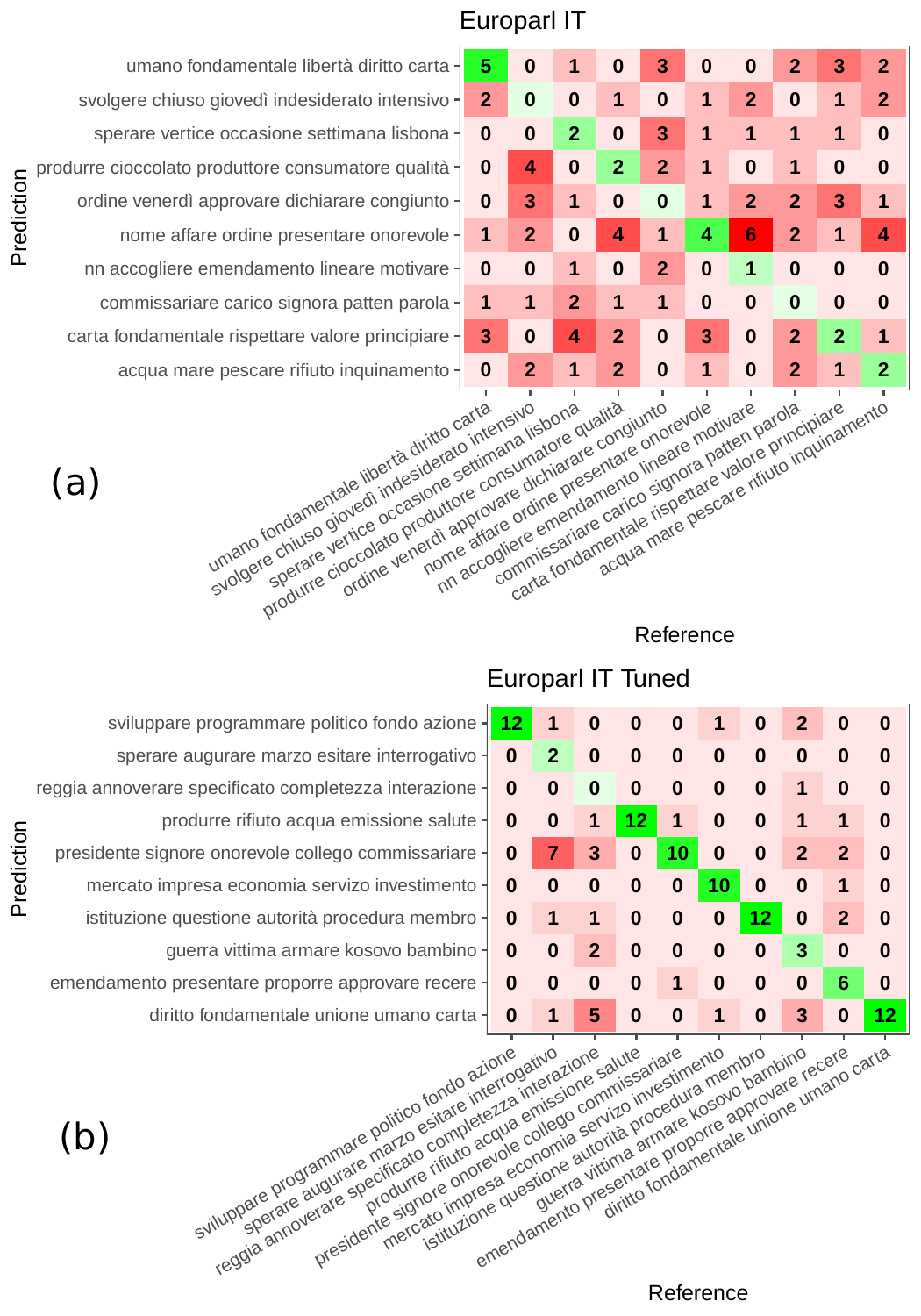}
\caption{Confusion matrices obtained by predicting the most likely topic from conditionally-generated sentences produced by ETC-NLG on the EuroParl Italian corpus. We leverage the same contextualized model that was used to annotate the unlabeled corpora to generate topic predictions and compare those against conditioning labels for sentences generated with weak conditioning parameters \lstinline{step_size}$=0.02$, \lstinline{gm_scale}$=0.9$ in (a) and strong conditioning parameters \lstinline{step_size}$=0.3$, \lstinline{gm_scale}$=0.95$ in (b). Rows represent topic predictions, columns are label references.}
\label{fig:cm_europarl_ita}
\end{figure}

Our parameter search experiments suggest that hyper-parameter tuning significantly influences the system's ability to generate conditioned text reliably.
In particular, stronger conditioning is achieved by setting higher \lstinline{step_size} and \lstinline{gm_scale} values w.r.t. their default settings in the original implementation of PPLMs~\cite{plugandplay}. The \lstinline{temperature} parameter, instead, appears to have a very moderate effect on conditioning.
Figures ~\ref{fig:cm_svevo},~\ref{fig:cm_europarl_eng},~\ref{fig:cm_europarl_ita} show the results obtained by imposing both a weaker conditioning (a) and a stronger conditioning (b), while maintaining fluency of the generated text in both cases. 
We obtain more coherent and diverse annotations on the Europarl corpora, thanks to the higher number of available samples. On the Svevo corpus, instead, we can notice the presence of ambiguous topics or repeated words among distinct annotations. We observe a higher fluency of text on weakly conditioned sentences (a) compared to strong conditioned sentences (b), which exhibit a higher topic coherence. The quality of the generated sentences can also be evaluated numerically from the confusion matrices, by computing the percentage of topic predictions which agree with the conditioning ones.
Training hyperparameters are reported in Table \ref{table:pplm_training} for reproducibility purposes. A closer manual inspection of the confusion matrices suggests that most misclassifications occur on topics that appear to us as ill-defined, confirming the importance of proper topic modeling for better automatic evaluation.
 
%% file: ijcol/sections/Conclusions.tex
\newpage
\section{Conclusions}

In this work, we presented ETC-NLG, an end-to-end method leveraging topic modeling annotations on unlabeled text corpora to generate topic-conditioned sentences in natural language. We highlighted this framework's strengths and weaknesses for English and Italian languages, mainly focusing on the more challenging scenario of the dialectal and archaic Italian language. We performed a thorough analysis of both generation and topic modeling performances. We concluded by presenting an experimental method to automatically evaluate the effectiveness of conditioning in the generated samples. Controlling the context of generated language is crucial for any real-world application involving human-machine interaction. Automating the evaluation procedure of generation models has the potential to improve their usability in realistic settings significantly. Our method aims at dealing with insufficient labeled training data and can be used to produce high quality conditioned text when provided with suitable topic models and parameters that balance generation fluency and conditioning strength.

Future developments of our approach should focus on two main bottlenecks in the ETC-NLG pipeline: developing better and more robust topic models for the automatic annotation of low-resource text and dealing with the computationally-heavy generation of conditioned sentences.

\begin{acknowledgments}
We would like to thank the anonymous reviewers for their insightful comments. This work was partly supported by a SISSA scholarship for Data Science students.
\end{acknowledgments}